\documentclass[runningheads]{llncs}
\usepackage[T1]{fontenc}
\usepackage[american]{babel}
\usepackage{graphicx}
\usepackage{booktabs}
\usepackage{multirow}
\usepackage{microtype}
\usepackage{amsmath}
\usepackage{amssymb}
\usepackage{amsfonts}
\usepackage{algorithm}
\usepackage{algpseudocode}
\usepackage{xspace}
\usepackage{orcidlink}
\usepackage{bbding}
\usepackage{hyperref}
\newcommand{\approach}{\textsc{LitEm}\xspace}
\newcommand{\approachbase}{\textsc{LitEm}\ensuremath{_{base}}\xspace}

\newcommand{\triple}[3]{\ensuremath{(#1, #2, #3)}}

\newcommand{\entities}{\ensuremath{\mathcal{E}}\xspace}
\newcommand{\relations}{\ensuremath{\mathcal{R}}\xspace}
\newcommand{\attributes}{\ensuremath{\mathcal{A}}\xspace}

\newcommand{\kg}{\ensuremath{\mathcal{G}}\xspace}
\newcommand{\scoreFunc}{\phi}
\begin{document}

\author{%
Rupesh Sapkota \orcidlink{0009-0009-3928-1130}(\Envelope) \and Louis Mozart Kamdem Teyou \orcidlink{0000-0001-7975-8794} \and Moshood Yekini \orcidlink{0009-0001-8120-3170}  \and  Caglar Demir\orcidlink{0000-0001-8970-3850} \and
Axel-Cyrille Ngonga Ngomo\orcidlink{0000-0001-7112-3516}
}

\institute{%
Data Science Group, Heinz Nixdorf Institute, Paderborn University, Germany\\
\email{\{rupesh.sapkota@,louis.mozart.kamdem.teyou@ moshood.olawale.yekini@, caglar.demir@, axel.ngonga@\}upb.de} 
}

\title{Neural Regression with Embeddings for Numerical Attribute Prediction in Knowledge Graphs}

\authorrunning{R. Sapkota et al.}
\titlerunning{Neural Regression with Embeddings}
\maketitle

\begin{abstract}
In recent years, transductive knowledge graph embedding models have been applied to tasks such as link prediction and query answering.
Although knowledge graphs often contain rich numerical attributes, most embedding models neglect them,
limiting their ability to represent real-world  knowledge graphs with diverse information.
In this work, we propose a neural regression model (\approach)
that enables transductive knowledge graph embedding models to predict numerical attributes within knowledge graphs.
Experimental results demonstrate that \approach achieves the best or second-best results on most attributes across FB15K-237, YAGO15K, DB15K, and Mutagenesis.  Furthermore, we propose a co-training framework that jointly trains state-of-the-art transductive knowledge graph embedding models with \approach, which improves link prediction performance mainly for bilinear models and simultaneously enables them to predict numerical attributes.
In addition, the literal-awareness evaluation demonstrates that co-training helps models to encode and exploit attribute information in a ``literal-aware'' manner, suggesting that the observed gains are not merely due to additional parameters. We publicly release our implementation at \url{https://github.com/dice-group/dice-embeddings}.

\keywords{Knowledge Graphs \and Embeddings \and Attribute Prediction}
\end{abstract}

\section{Introduction}
\label{section:introduction}

 A Knowledge Graph (KG)  represents real-world entities and their relationships in a structured manner, enabling knowledge representation, information retrieval, and decision-making across domains such as search engines, drug discovery, natural language processing and recommendation systems \cite{zou2020survey}.  Publicly available general-purpose KGs such as DBpedia \cite{lehmann2015dbpedia}, Freebase \cite{bollacker2008freebase}, and YAGO \cite{suchanek2007yago} store large-scale real-world knowledge about entities and their relationships. However, many existing KGs suffer from incompleteness, which limits their ability to support reasoning tasks  due to missing facts \cite{min2013distant}.

Link prediction is widely used to address incompleteness in KGs by inferring missing links between entities \cite{bordes2013translating,trouillon2016complex}. Knowledge Graph Embedding (KGE) models represent entities and relations in vector spaces and perform link prediction with state-of-the-art performance. Numerous KGE methods have been proposed in the literature \cite{bordes2013translating,yang2014embedding,trouillon2016complex,sun2019rotate,demir2021convolutional,demir2023clifford}, but they rely solely on relational information between entities and overlook literals (textual descriptions, date-time values, numerical values, measurements, images, and other attribute data) that can enrich entity representations \cite{gesese2021survey}. In particular, numerical literals have been widely used to improve link prediction performance \cite{tay2017multi,wu2018knowledge,kristiadi2019incorporating,klironomos2025realite} and other tasks such as complex query answering \cite{demir2023litcqd}. Such methods either modify the scoring function or incorporate literals during training for numerical attribute imputation and link prediction. Prior work suggests pretrained KGEs already encode much of the information needed to predict numeric attributes, without modifying the embedding structures~\cite{xue2022introducing}. We build on this with an 
approach that learns attribute representations from pretrained embeddings and requires no changes to the KGE scoring function, enabling it to both predict and augment numerical attributes for any transductive KGE model.
Our approach, model-agnostic in design, can be paired with existing KGE models without modifying their scoring functions in transductive settings to simultaneously augment and predict numerical attributes. In this work, we provide the following contributions:
\begin{itemize}
    \item Propose \approach, a neural regression model that predicts numerical attributes from pre-trained embeddings.
\item Design a framework that co-trains KGE models with \approach to jointly predict and augment numerical attributes.
\item Introduce a gradient-based dynamic weighting mechanism to balance KGE and literal losses.
\item Repurpose DB15K and Mutagenesis as benchmarks for numerical literal prediction, with the first reported results on both.
\end{itemize}

\section{Background}
\label{section:background}
\subsection{Knowledge Graphs}

A KG is a structured network representing real-world entities as nodes and their relationships as edges, conveying information in the form of triples \cite{hogan2021knowledge}. In addition to relational triples, KGs also contain triples that associate entities with their attribute values. While many definitions of KGs exist, we consider a KG defined as
$\kg := \{ \triple{h}{r}{t} \} \cup \{ \triple{e}{a}{v} \} \subseteq \entities \times \relations \times \entities \;\cup\; \entities \times \attributes \times \mathcal{V} $
where \(\entities\), \(\relations\), and \(\attributes\) denote the sets of entities, relations, and attributes, respectively, and $\mathcal{V} \subset \mathbb{R}$ represents a subset of real numbers
~\cite{wang2022augmenting,kristiadi2019incorporating,demir2023litcqd}.  A relational triple  is represented as \(\triple{h}{r}{t}\), where \(h, t \in \entities\) and \(r \in \relations\). Similarly, an attribute (numerical) triple is represented as \(\triple{e}{a}{v}\), where \(e \in \entities\), \(a \in \attributes\) denotes a numerical attribute,  and \(v \in \mathcal{V} \). For example, \triple{Berlin}{CapitalOf}{Germany} is a relational triple, while \triple{Berlin}{hasPopulation}{3{,}960{,}000} is an example of an attribute (literal) triple.

\subsection{Transductive Knowledge Graph Embeddings}
KGE models represent entities and relations in vector spaces, facilitating downstream tasks such as link prediction~\cite{demir2023clifford} and neural reasoning~\cite{demir2023litcqd}.
KGE models are often formulated using parameterized scoring functions of the form
$\scoreFunc_\Theta: \entities \times \relations \times \entities \to \mathbb{R},$
where the parameters \(\Theta\) typically include embeddings for entities \(\mathbf{E} \in \mathbb{V}^{|\mathcal{E}| \times d_e}\) and relations \(\mathbf{R} \in \mathbb{V}^{|\mathcal{R}| \times d_r}\), along with other learnable components \cite{demir2023clifford}, where $\mathbb{V}$ denotes $d$-dimensional vector space and $d \in \mathbb{N},  d > 0$; here entities and relations share the same embedding dimensionality \textit{d}.  Embedding methods vary in how they represent relations: TransE~\cite{bordes2013translating} as additive translations, DistMult~\cite{yang2014embedding} via a multiplicative scoring function, ComplEx~\cite{trouillon2016complex} in complex-valued space to capture asymmetry, RotatE~\cite{sun2019rotate} as rotations in complex space, QMult and OMult~\cite{demir2021convolutional} as quaternion and octonion extensions of DistMult and ComplEx, DualE~\cite{cao2021dual} via dual quaternions, and Keci~\cite{demir2023clifford} in Clifford algebra.

\subsection{Literal Embedding Models}

Literal embedding models extend KGE approaches by learning embeddings for numerical attributes, capturing real-valued properties of entities in the representation space. Most existing approaches, however, do not leverage transductive KGE models to jointly learn attribute representations and predict numerical values.
Let $a \in \attributes$ denote a numerical attribute of an entity $e \in \entities$, and let $v \in \mathcal{V}$ represent its associated numerical value. A literal embedding model defines a scoring function $\phi'_{\Theta'} : \entities \times \attributes \rightarrow \mathbb{R}$, where the parameters $\Theta'$ typically include attribute embeddings $\mathbf{A} \in \mathbb{R}^{|\mathcal{A}| \times d_a}$, and $d_a$ denotes the dimensionality of literal embeddings. The predicted numerical value is given by $\hat{v} = \phi'_{\Theta'}(e,a)$. 

\section{Related Work}
\label{section:related_work}
Prior work on numerical attributes in knowledge graphs can be broadly grouped into attribute prediction and literal-aware embedding approaches. For numerical attribute prediction, Tay et al.~\cite{tay2017multi} jointly learned relational and attribute networks in a shared embedding space, Kotnis and García-Durán~\cite{kotnis2019learning} proposed regression- and propagation-based baselines (LR, NAP, LR++, NAP++, GLOBAL, LOCAL), Bayram et al.~\cite{bayram2021node} introduced MrAP, which propagates attribute values over multi-relational neighbourhoods, and Xue et al.~\cite{xue2022introducing} combined graph-based methods with a pre-trained language model for masked numerical value prediction (KGE-Reg). Other approaches incorporate literals directly into KGE models: TransEA~\cite{wu2018knowledge} jointly optimises relational and attribute embeddings, KBLRN~\cite{Garcia-DuranN18} combines relational, latent, and numerical features in a product-of-experts framework, LiteralE~\cite{kristiadi2019incorporating} injects literals into entity embeddings through a learnable gate, KGA discretises literals into numerical bins linked to entities, exposing them to standard KGE models~\cite{wang2022augmenting}, and ReaLitE~\cite{klironomos2025realite} enriches relation embeddings with literal information for link prediction. \approach\ differs from these approaches in what it modifies. Where KGE-Reg predicts masked values over a language model, \approach\ regresses directly from frozen entity embeddings through a gated-residual network, and unlike Tay et al.'s multi-task setup it adds no auxiliary objective to the KGE model. LiteralE also uses a gate, but applies it to entity embeddings during KGE training, whereas \approach\ gates the prediction of attributes from embeddings that already exist. \approach\ then pairs this decoder with model-agnostic co-training that updates those embeddings through the literal loss alone, leaving the KGE scoring function unchanged.

\section{Literal Embedding Model}
\label{section:approach}
In this section, we present \textbf{Lit}eral \textbf{Em}bedding (\textbf{\approach}), a neural embedding regression model that learns representations and predicts numerical attributes associated with entities in a knowledge graph. The model uses the pre-trained embeddings of entities as input features and performs regression to predict the numerical values of different attributes of entities present in the knowledge graph. Given an attribute triple $\triple{e}{a}{v} \in \kg$, where $e \in \mathcal{E}$ and $ a \in \mathcal{A}$, the model takes the entity-attribute pair $(\textit{e},\textit{a})$ as input and is trained to  predict $\textit{v}$, the  associated numerical value  using the regression function defined in Equation ~\ref{eq:litem_base}. 

\begin{equation}
\label{eq:litem_base}
\approach_{base}(\textit{e},\textit{a}) = \mathbf{W}_2 \cdot \left( \mathbf{ReLU}(\mathbf{W}_1 \cdot [\textbf{e},\textbf{a}] + \mathbf{b}_1) + [\textbf{e},\textbf{a}] \right) + \mathbf{b}_2.
\end{equation}

We refer to Equation~\ref{eq:litem_base} as the base model, where  $\mathbf{e} \in \textbf{E}$ and $\mathbf{a} \in \textbf{A}$ denote the embeddings of the entity $e$ and attribute $a$, respectively, and share the same embedding dimensionality $d$. In this formulation, the attribute embedding $\mathbf{a}$ is a trainable parameter, whereas the entity embeddings $\mathbf{e}$, obtained from a pre-trained KGE model and fixed by default (optionally can be updated during \approach training). The model first concatenates the entity and attribute embeddings $[\mathbf{e}, \mathbf{a}]$, applies a linear transformation ($\mathbf{W}_1 \in \mathbb{R}^{2d \times 2d}$, $\mathbf{b}_1 \in \mathbb{R}^{2d}$) followed by a ReLU activation, and adds the original input via a residual connection. The resulting representation is then projected to a scalar output through a second linear layer ($\mathbf{W}_2 \in \mathbb{R}^{1 \times 2d}$, $\mathbf{b}_2 \in \mathbb{R}$). To selectively incorporate literal information, we further extend the model with a \emph{Gated Residual} mechanism inspired by \cite{kristiadi2019incorporating}, allowing the model to control the contribution of the residual connection. The complete $\approach(\mathbf{e},\mathbf{a})$ model is defined in Equation~\ref{eq:litem}.

\begin{align}
\mathbf{z} &= \mathbf{ReLU}(\mathbf{W}_1 \cdot [\mathbf{e}, \mathbf{a}] + \mathbf{b}_1), \nonumber \quad 
\begin{bmatrix} \mathbf{u} \\ \mathbf{g} \end{bmatrix} &= \mathbf{W}_{\text{res}} \cdot [\mathbf{z} , [\mathbf{e}, \mathbf{a}]] + \mathbf{b}_{\text{res}} \nonumber\\
\quad \approach(\mathbf{e},\mathbf{a}) &= \mathbf{W}_2 \cdot \left( \mathbf{u} \odot \sigma(\mathbf{g}) \right) + \mathbf{b}_2,
\label{eq:litem}
\end{align}

The gating mechanism projects the concatenated vector $[\mathbf{z}, [\mathbf{e}, \mathbf{a}]]$ through the parameters $\mathbf{W}_\text{res} \in \mathbb{R}^{4d \times 4d}$ and $\mathbf{b}_\text{res} \in \mathbb{R}^{4d}$( as $[\mathbf{e}, \mathbf{a}] $ and $\mathbf{z}\in\mathbb{R}^{2d}$ ), splits the result into a value $\mathbf{u}$ and a gate $\mathbf{g}$. The value is then modulated via element-wise multiplication, denoted by $\odot$, where $\sigma$ denotes the sigmoid activation function.

\subsection{Incorporating Literal Information into KGEs}
\begin{algorithm}[!t]
\footnotesize
\caption{Combined Training of Embedding Model with \approach}
\label{alg:combined}
\begin{algorithmic}[1]
\State \textbf{Input:} Triples $\mathcal{G} = \{(h, r, t)\} \cup \{(e, a, v)\}$,labels $q \in \{0,1\}$ for relational triples; learning rates $\gamma_{KGE}$, $\gamma_{\approach}$, epochs $n$, batch size $b$, models \textbf{KGE}, \textbf{\approach}
\State \textbf{Initialize:} $\theta = \{\theta_{KGE}, \theta_{\approach}\}$
\For{epoch $= 1$ \textbf{to} $n$}
    \For{mini-batch $B = \{(h, r, t)\}$ of size $b$ }
        \State $\mathbf{h}, \mathbf{r}, \mathbf{t} \gets \textbf{KGE}_{\theta}(h, r, t)$
        \State $\hat{q} \gets \sigma(\phi_\theta(h, r, t))$
        \State $\mathcal{L}_{KGE} \gets -\frac{1}{b} \sum [q \log(\hat{q}) + (1-q)\log(1-\hat{q})]$
        \State Sample : $B_a$ $\gets$ \(\{(e, a, v) \in \mathcal{G} \mid h = e\}\) \Comment{Attribute triples for batch}
        \State $\hat{y} \gets \textbf{LitEm}_{\theta}(e, a)$
        \State $\mathcal{L}_{\approach} \gets \frac{1}{|B_a|} \sum |\hat{y} - v|$
        \State $\lambda \gets \min\left(\dfrac{\lVert \nabla_{\mathbf{E}} \mathcal{L}_{KGE} \rVert_2}{\lVert \nabla_{\mathbf{E}} \mathcal{L}_{\approach} \rVert_2 + \varepsilon},\, w_{\approach}\right)$ \Comment{dynamic mixing weight; $w_{\approach}=0.5$, $\varepsilon=10^{-12}$}
        \State $\mathcal{L}_{\text{combined}} \gets (1-\lambda)\mathcal{L}_{KGE} + \lambda \mathcal{L}_{\approach}$
        \State $\theta_{KGE} \gets \theta_{KGE} - \gamma_{KGE} \nabla_{\theta_{KGE}} \mathcal{L}_{\text{combined}}$
        \State $\theta_{\approach} \gets \theta_{\approach} - \gamma_{\approach} \nabla_{\theta_{\approach}} \mathcal{L}_{\text{combined}}$
    \EndFor
\EndFor
\State \Return $\theta$
\end{algorithmic}
\end{algorithm}

Building on the \approach model, we introduce a co-training framework that jointly trains any transductive KGE model with \approach (Algorithm~\ref{alg:combined}), injecting attribute information directly into the entity embeddings.
During training, each relational triple $\triple{h}{r}{t}$ is paired with the numerical attributes of its head entities as literal triples $\triple{e}{a}{v}$. The KGE model processes the relational triple while \approach processes the attribute triples. Entity embeddings stay fixed in the standalone \approach, but the combined framework updates them through the literal prediction loss, so both structural and numerical signals shape the shared embeddings.

 Rather than a fixed loss weight, our co-training approach computes a dynamic scaling factor $\lambda$ from the relative gradient norms of the KGE loss and the \approach loss on the shared entity embeddings. The adaptive $\lambda$ grows when the structural gradient dominates, upweighting the literal loss so its signal keeps pace, and shrinks when the literal gradient is large. Capping it at $0.5$ guarantees $1 - \lambda \geq 0.5$, so the structural objective always retains at least half the weight.

\section{Evaluation}
\label{section:evaluation}

\subsubsection{Experimental Setup.}
All experiments were carried out in a Python 3.10 environment with GPU acceleration, and KGE models are implemented using the DICE-Embeddings framework \cite{demir2022hardware}. For the literal prediction task, we follow \cite{kotnis2019learning} and use TransE embeddings trained on all relational triples from the training, validation, and test set to ensure a fair comparison between \approach and the baselines. Although KGE-Regression \cite{xue2022introducing} reports results from an ensemble of embedding models, we include only its TransE-based results for consistency. For link prediction and co-training, all KGE models use 64-dimensional real-valued entity and relation embeddings.
Hyperparameters are selected on the validation split and kept fixed across test evaluations. Standalone \approach is trained with MAE loss and Adam on frozen entity embeddings, using a batch size of 256, with the number of epochs (160--280), the learning rate (0.0005--0.01), and the dropout rate (0.1--0.3) tuned per dataset. The KGE and co-training runs use BCE loss, Adam, and KvsAll scoring, with 300 epochs, a batch size of 1024, and a learning rate of 0.05 applied uniformly across all models. Supplementary materials, detailed hyperparameters, and trained models are available online.\footnote{\url{https://github.com/dice-group/literal-embeddings/}}

\subsubsection{Model Complexity.} Let $d$ be the embedding dimension and $|\mathcal{A}|$ the number of numerical attributes. For the base \approach variant without the residual component, the  trainable parameters are $|\mathcal{A}|d + 4d^2 + 4d + 1$, while for the standard \approach architecture they are $|\mathcal{A}|d + 20d^2 + 8d + 1$. Thus, the parameter complexity is linear in the number of attributes and quadratic in the embedding dimension, while the per-instance computational cost of the literal head remains $O(d^2)$ for both training and inference. In the combined training setting, the only additional parameters come from the literal embedding model.

\begin{table}[t]
\caption{Dataset statistics for relational and literal triples including train, test and validation splits. The $|R|/|A|$ column reports the number of relations for relational datasets and numerical attributes for literal datasets.}
    \centering
    \footnotesize
    \setlength{\tabcolsep}{3pt}
    \label{tab:datasets}
    \begin{tabular}{@{}lllllll@{}}
 \toprule
 \textbf{Dataset} & \textbf{Type} & $|\mathcal{E}|$ & $|\mathcal{R}|$ / $|\mathcal{A}|$ & $|\mathcal{G}^{\text{Train}}|$ & $|\mathcal{G}^{\text{Val}}|$ & $|\mathcal{G}^{\text{Test}}|$ \\
 \midrule

 \multirow{2}{*}{DB15K} 
 & Relational & 12842 & 279 & 79222 & 9903 & 9903 \\
 & Literal    & 7078  & 9   & 9464  & 1183 & 1184 \\
 
 \midrule
 \multirow{2}{*}{FB15K-237} 
 & Relational & 14541 & 237 & 272115 & 20466 & 17535 \\
 & Literal    & 10066 & 11  & 18615  & 2329  & 2306 \\
  \midrule
 \multirow{2}{*}{Mutagenesis} 
 & Relational & 14250 & 8   & 33281  & 16239 & 5503 \\
 & Literal    & 6124  & 4   & 3950    & 659 & 1975\\
 \midrule
 \multirow{2}{*}{YAGO15K} 
 & Relational & 15403 & 32  & 110441 & 13800 & 13815 \\
 & Literal    & 15081 & 7   & 18825  & 2354  & 2353 \\

       \bottomrule
    \end{tabular}
\end{table}

\subsubsection{Datasets.}Table~\ref{tab:datasets} summarizes the four datasets: FB15K-237, YAGO15K, DB15K, and Mutagenesis. FB15K-237 is a subset of Freebase \cite{bollacker2008freebase}, while YAGO15K and DB15K are derived from YAGO and DBpedia with entities aligned to FB15K \cite{liu2019mmkg}, with numerical literals taken from \cite{xue2022introducing}. Mutagenesis \cite{westphal2019sml} is an RDF KG for class expression learning and numerical attribute prediction. Since numerical attributes vary widely in scale, we apply z-normalization following \cite{kotnis2019learning} for standalone literal prediction and min-max normalization for combined training.

\subsection{Results}
\begin{table}[t]
\caption{Literal prediction performance of \approachbase and \approach against various approaches, evaluated using Mean Absolute Error (MAE) over 5 runs on FB15K-237 and YAGO15K. Lower MAE is better; best and second-best means are bolded and underlined, respectively. KGA, NAP++ and MrAP results are taken from \cite{wang2022augmenting}; the performance of LR on YAGO15K is not reported in \cite{kotnis2019learning}.}
    \centering
    \scriptsize
    \setlength{\tabcolsep}{2pt}
    \label{tab:lit_pred_fb_yago}
\begin{tabular}{@{}lrrrrrrr@{}}
\toprule
\textbf{Relation} & \textbf{LR} & \textbf{KGE-reg} & \textbf{KGA} & \textbf{NAP++} & \textbf{MrAP} & \textbf{\approachbase} & \textbf{\approach}\\
\midrule
\multicolumn{8}{c}{\textbf{FB15K-237}} \\
\midrule
release\_date & 5.590 & 5.526 & \underline{4.000} & 9.900 & 6.300 & 4.181 & \textbf{3.449} \\
loc.date\_founded & 145.460 & 153.387 & \textbf{76.000} & \underline{92.100} & 98.800 & 153.466 & 149.237 \\
latitude & 8.470 & 8.741 & \underline{2.100} & 11.800 & \textbf{1.500} & 3.577 & 2.596 \\
longitude & 25.560 & 24.959 & 7.100 & 54.700 & \textbf{4.000} & 9.517 & \underline{4.424} \\
loc.area & 7.70e+05 & 1.78e+06 & \textbf{6.10e+04} & \underline{4.40e+05} & \underline{4.40e+05} & 1.53e+06 & 1.37e+06 \\
org.date\_founded & 53.850 & 71.632 & 49.000 & 59.300 & 58.300 & \underline{48.598} & \textbf{47.762} \\
date\_of\_death & 35.890 & 45.575 & 20.600 & 52.300 & \underline{16.300} & 21.286 & \textbf{15.752} \\
date\_of\_birth & 26.600 & 23.489 & 18.900 & 22.100 & \underline{15.000} & 15.076 & \textbf{11.935} \\
height\_meters & \textbf{0.065} & 0.162 & 0.077 & 0.080 & 0.086 & 0.077 & \underline{0.076} \\
weight\_kg & - & \underline{9.515} & 11.600 & 15.300 & 12.900 & 11.400 & \textbf{9.442} \\
pop.\_number & 7.90e+06 & 2.60e+07 & 4.00e+06 & 7.50e+06 & 2.10e+07 & \underline{2.94e+06} & \textbf{2.33e+06} \\

\midrule
\multicolumn{8}{c}{\textbf{YAGO15K}} \\
\midrule
diedOnDate & - & 50.119 & \underline{30.800} & 45.700 & 34.000 & 35.680 & \textbf{30.135} \\
happenedOnDate & - & 96.691 & \underline{29.900} & 73.700 & 54.100 & \textbf{26.842} & 28.260 \\
Latitude & - & 8.656 & 3.400 & 8.700 & 2.800 & \underline{2.456} & \textbf{1.703} \\
Longitude & - & 25.907 & 7.200 & 43.100 & \underline{5.700} & 9.518 & \textbf{5.598} \\
BornOnDate & - & 19.795 & 16.300 & 23.200 & 19.700 & \underline{13.939} & \textbf{13.242} \\
CreatedOnDate & - & 70.586 & 58.200 & 83.500 & 70.400 & \textbf{56.267} & \underline{57.206} \\
DestroyedOnDate & - & 41.972 & 23.300 & 38.200 & 34.600 & \underline{22.379} & \textbf{21.673} \\
\bottomrule
\end{tabular}

\end{table}

\begin{table}[t]
\caption{Literal prediction performance of different embedding models via \approach (averaged over 5 runs) on the FB15K-237 and YAGO15K datasets, evaluated using Mean Absolute Error (MAE).}
    \centering
    \scriptsize
    \setlength{\tabcolsep}{2pt}
    \begin{tabular}{lrrrrrrrr}
\toprule
    \textbf{Relation} & \textbf{ComplEx} &  \textbf{DistMult} & \textbf{DualE} &  \textbf{Keci} &  \textbf{OMult} & \textbf{QMult} & \textbf{RotatE} & \textbf{TransE}\\
    \midrule
    \multicolumn{8}{c}{\textbf{FB15K-237}} \\
     \midrule
release\_date & 4.863 & 5.000 & 5.283 & 5.012 & 5.052 & 5.075 & 4.355 & \textbf{3.449} \\
loc.date\_founded & 198.301 & 204.629 & 205.871 & 193.153 & 175.986 & 206.065 & \textbf{148.162} & 149.237 \\
latitude & 7.859 & 7.884 & 7.274 & 7.357 & 7.390 & 6.813 & 2.882 & \textbf{2.596} \\
longitude & 18.484 & 19.141 & 17.994 & 16.236 & 18.422 & 17.858 & 5.003 & \textbf{4.424} \\
org.date\_founded & 63.544 & 61.424 & 61.757 & 63.811 & 62.953 & 63.361 & \textbf{46.298} & 47.762 \\
date\_of\_death & 23.664 & 26.049 & 25.161 & 26.615 & 24.060 & 22.788 & 15.825 & \textbf{15.752} \\
date\_of\_birth & 18.235 & 17.724 & 17.536 & 18.760 & 18.264 & 18.641 & 13.194 & \textbf{11.935} \\
height\_meters & \textbf{0.068} & 0.070 & 0.070 & 0.069 & 0.072 & 0.070 & 0.074 & 0.076 \\
weight\_kg & 10.308 & 9.573 & 9.132 & 11.476 & 9.606 & 9.879 & \textbf{8.029} & 9.442 \\
    \midrule
    \multicolumn{8}{c}{\textbf{YAGO15K}} \\
    \midrule
diedOnDate & 29.418 & 32.797 & \textbf{28.583} & 38.620 & 28.958 & 30.261 & 29.620 & 30.135 \\
happenedOnDate & 39.133 & 37.490 & 43.162 & 40.985 & 36.342 & 39.913 & 28.618 & \textbf{28.260} \\
Latitude & 5.672 & 5.502 & 5.338 & 6.403 & 5.923 & 6.163 & 2.030 & \textbf{1.703} \\
Longitude & 28.787 & 25.142 & 24.623 & 28.572 & 24.449 & 27.378 & \textbf{4.761} & 5.598 \\
BornOnDate & 19.339 & 20.063 & 17.090 & 19.992 & 16.811 & 18.547 & 14.354 & \textbf{13.242} \\
CreatedOnDate & 68.848 & 70.488 & 66.652 & 70.660 & 66.099 & 70.231 & \textbf{55.887} & 57.206 \\
DestroyedOnDate & 27.655 & 31.172 & 28.288 & 32.285 & 25.405 & 26.363 & 23.979 & \textbf{21.673} \\
\bottomrule
\end{tabular}
    
    \label{tab:kge_litpreds}
\end{table}

Table~\ref{tab:lit_pred_fb_yago} compares \approach\ with baseline and state-of-the-art approaches on FB15K-237 and YAGO15K using Mean Absolute Error (MAE). On FB15K-237, \approach\ outperforms the embedding-based regression baseline KGE-Reg on all attributes and LR on most attributes. On YAGO15K, \approach\ and \approachbase achieve the strongest performance across all attributes, with consistent gains on both temporal and spatial attributes. The base model remains competitive on several attributes, indicating that the entity-attribute interaction provides a useful signal, which the gated residual mechanism further refines. MrAP outperforms \approach on spatial attributes in FB15K-237, likely because neighbourhood propagation exploits geographic locality among connected entities, whereas on YAGO15K \approach\ decodes spatial signals from the embeddings well enough to outperform MrAP on both latitude and longitude. Counting both variants, \approach\ or \approachbase ranks best or second-best on 8 of the 11 FB15K-237 attributes and on all 7 YAGO15K attributes. The weaker performance on attributes such as \textit{loc.date\_founded} and \textit{loc.area} can be attributed to \approach's shared multi-attribute regression setting, which makes extreme values in low-density tails harder to estimate.

Similarly, Table~\ref{tab:kge_litpreds} reveals a consistent pattern on FB15K-237 and YAGO15K: translational models such as TransE and RotatE achieve lower literal prediction loss than models with multiplicative scoring functions.  TransE obtains the best MAE for most attributes on both datasets, while RotatE performs similarly on most attributes. This likely reflects that translational geometries place semantically related entities closer together in embedding space, making their numerical attributes easier to recover via regression. In contrast, models with more complex scoring functions may require scoring-function-aware literal decoders to better exploit the attribute signals captured in their embeddings.

\begin{table}[t]
\caption{Literal prediction performance of baseline approaches and \approach (averaged over 5 runs) on the DB15K and Mutagenesis datasets, evaluated using Mean Absolute Error (MAE).}
\label{tab:lit_pred_db15k}
\centering
\footnotesize
\setlength{\tabcolsep}{2pt}
\begin{tabular}{@{}lcccccc@{}}
\toprule
\textbf{Relation} & \textbf{LOCAL} & \textbf{GLOBAL} & \textbf{Lin-Reg} & \textbf{MrAP} & \textbf{\approachbase} & \textbf{\approach}\\
\midrule
\multicolumn{7}{c}{\textbf{DB15K}} \\
\midrule
birthDate & 23.361 & 21.695 & 16.702 & 18.189 & \textbf{12.690} & 13.212 \\
completionDate & 9.425 & 7.692 & 23.895 & \textbf{3.942} & 5.454 & 4.999 \\
deathDate & 26.636 & 25.987 & 24.031 & 17.527 & \textbf{15.430} & 15.825 \\
formationDate & 40.358 & 40.358 & 77.358 & \textbf{32.234} & 40.715 & 38.772 \\
foundingDate & 52.944 & 52.407 & 40.059 & 44.478 & \textbf{35.551} & 40.270 \\
height & 1.403 & 1.403 & 7.592 & 1.403 & 0.502 & \textbf{0.346} \\
releaseDate & 13.332 & 11.583 & 12.500 & 9.156 & \textbf{8.352} & 9.213 \\
latitude & 11.830 & 7.081 & 7.664 & 6.298 & 4.235 & \textbf{3.758} \\
longitude & 59.821 & 29.056 & 25.577 & 16.992 & 14.826 & \textbf{11.776} \\
\midrule
\multicolumn{7}{c}{\textbf{Mutagenesis}} \\
\midrule
mutagenesis\#act & 1.989 & 1.989 & 3.100 & 1.902 & 1.434 & \textbf{1.348} \\
mutagenesis\#charge & 0.197 & 0.197 & 0.139 & 0.397 & 0.030 & \textbf{0.022} \\
mutagenesis\#logp & 1.299 & 1.299 & 1.452 & 1.243 & 0.753 & \textbf{0.655} \\
mutagenesis\#lumo & 0.365 & 0.365 & 1.132 & 0.445 & 0.280 & \textbf{0.248} \\
\bottomrule
\end{tabular}

\end{table}

Table~\ref{tab:lit_pred_db15k} reports the performance of \approach and baseline approaches on DB15K and Mutagenesis datasets. Since DB15K and Mutagenesis are introduced in this work as new benchmarks for numeric attribute prediction, no prior approaches report results on them, so we adopt per-attribute linear regression as a standard baseline and additionally implement MrAP. \approach\ variants outperform the linear regression baseline across all attributes in both datasets and outperform MrAP on the majority of attributes, with particularly notable gains on \textit{longitude} (11.776 vs. 16.992), \textit{height} (0.346 vs. 1.403), and \textit{mutagenesis\#charge} (0.022 vs. 0.397). Counting both variants, \approach\ or \approachbase ranks best or second-best on all 9 DB15K and all 4 Mutagenesis attributes.

\begin{table}[!ht]
    \caption{Link prediction performance of  embedding models with 64-dimensional real-valued vectors on the test sets of  FB15k-237, YAGO15K  and DB15K datasets for standard and combined training. The models are trained for 300 epochs with a batch size of 1024 and a learning rate of 0.05, and values in bold mark the best per-metric value within each model-dataset combination. $+\,$\approach denotes performance with combined training and $+\,$ReaLitE denotes performance of the ReaLitE model \cite{klironomos2025realite}.}
    \centering
    \scriptsize
    \label{tab:lp}
    \setlength{\tabcolsep}{2pt}
     \begin{tabular}{l cccc c cccc c cccc}
\toprule
\textbf{Models} & \multicolumn{4}{c}{\textbf{FB15K-237}} & &\multicolumn{4}{c}{\textbf{YAGO15K}} && \multicolumn{4}{c}{\textbf{DB15K}} \\
\cmidrule(l){2-5} \cmidrule(l){7-10} \cmidrule(l){12-15}
& MRR & @1 & @3 & @10 & &MRR & @1 & @3 & @10 & &MRR & @1 & @3 & @10 \\
\midrule
ComplEx & 0.189 & 0.127 & 0.204 & 0.313 &  & 0.217 & \textbf{0.159} & 0.236 & 0.334 &  & 0.284 & \textbf{0.235} & 0.303 & 0.377 \\
+ReaLitE & 0.169 & 0.111 & 0.182 & 0.286 &  & 0.204 & 0.139 & 0.228 & 0.332 &  & 0.270 & 0.221 & 0.290 & 0.356 \\
+\approach & \textbf{0.195} & \textbf{0.133} & \textbf{0.213} & \textbf{0.319} &  & \textbf{0.219} & 0.149 & \textbf{0.242} & \textbf{0.357} &  & \textbf{0.285} & 0.234 & \textbf{0.306} & \textbf{0.384} \\
\midrule
DistMult & 0.185 & 0.120 & 0.204 & 0.314 &  & 0.186 & 0.111 & 0.212 & 0.333 &  & 0.259 & 0.208 & 0.278 & 0.354 \\
+ReaLitE & 0.174 & 0.115 & 0.189 & 0.293 &  & 0.207 & 0.138 & 0.231 & 0.342 &  & \textbf{0.274} & \textbf{0.222} & \textbf{0.296} & 0.374 \\
+\approach & \textbf{0.204} & \textbf{0.142} & \textbf{0.220} & \textbf{0.326} &  & \textbf{0.240} & \textbf{0.167} & \textbf{0.264} & \textbf{0.383} &  & 0.271 & 0.212 & 0.294 & \textbf{0.383} \\
\midrule
DualE & 0.189 & 0.126 & 0.204 & 0.314 &  & 0.212 & 0.148 & 0.233 & 0.338 &  & 0.281 & 0.237 & 0.292 & 0.369 \\
+ReaLitE & 0.153 & 0.105 & 0.164 & 0.245 &  & \textbf{0.324} & \textbf{0.255} & \textbf{0.356} & \textbf{0.455} &  & 0.137 & 0.097 & 0.150 & 0.211 \\
+\approach & \textbf{0.199} & \textbf{0.139} & \textbf{0.216} & \textbf{0.319} &  & 0.225 & 0.151 & 0.252 & 0.374 &  & \textbf{0.285} & \textbf{0.239} & \textbf{0.299} & \textbf{0.376} \\
\midrule
Keci & 0.205 & 0.139 & 0.221 & 0.338 &  & 0.196 & 0.127 & 0.220 & 0.334 &  & 0.272 & 0.222 & 0.290 & 0.371 \\
+ReaLitE & 0.204 & 0.134 & 0.222 & 0.347 &  & 0.198 & 0.121 & 0.222 & \textbf{0.356} &  & 0.309 & 0.257 & 0.330 & 0.406 \\
+\approach & \textbf{0.239} & \textbf{0.167} & \textbf{0.260} & \textbf{0.383} &  & \textbf{0.206} & \textbf{0.130} & \textbf{0.231} & \textbf{0.356} &  & \textbf{0.320} & \textbf{0.262} & \textbf{0.346} & \textbf{0.425} \\
\midrule
OMult & 0.170 & 0.110 & 0.183 & 0.291 &  & 0.250 & 0.188 & 0.278 & 0.368 &  & 0.239 & 0.200 & 0.250 & 0.308 \\
+ReaLitE & 0.149 & 0.090 & 0.164 & 0.263 &  & 0.201 & 0.136 & 0.223 & 0.334 &  & 0.222 & 0.197 & 0.227 & 0.266 \\
+\approach & \textbf{0.179} & \textbf{0.115} & \textbf{0.196} & \textbf{0.305} &  & \textbf{0.262} & \textbf{0.193} & \textbf{0.292} & \textbf{0.393} &  & \textbf{0.281} & \textbf{0.238} & \textbf{0.295} & \textbf{0.366} \\
\midrule
QMult & 0.190 & 0.127 & 0.206 & 0.318 &  & 0.222 & 0.161 & 0.244 & 0.343 &  & 0.251 & 0.213 & 0.260 & 0.324 \\
+ReaLitE & 0.211 & 0.143 & 0.227 & 0.349 &  & 0.203 & 0.139 & 0.223 & 0.334 &  & \textbf{0.291} & \textbf{0.246} & \textbf{0.310} & \textbf{0.375} \\
+\approach & \textbf{0.219} & \textbf{0.154} & \textbf{0.235} & \textbf{0.351} &  & \textbf{0.237} & \textbf{0.170} & \textbf{0.263} & \textbf{0.370} &  & 0.270 & 0.222 & 0.287 & 0.364 \\
\midrule
RotatE & \textbf{0.327} & \textbf{0.238} & \textbf{0.360} & \textbf{0.506} &  & \textbf{0.355} & \textbf{0.272} & \textbf{0.389} & \textbf{0.514} &  & \textbf{0.385} & \textbf{0.302} & \textbf{0.428} & \textbf{0.539} \\
+ReaLitE & 0.066 & 0.036 & 0.070 & 0.124 &  & 0.331 & 0.249 & 0.363 & 0.489 &  & 0.244 & 0.182 & 0.270 & 0.368 \\
+\approach & 0.325 & 0.236 & 0.358 & 0.501 &  & 0.346 & 0.262 & 0.380 & 0.507 &  & 0.382 & 0.299 & 0.427 & 0.537 \\
\midrule
TransE & \textbf{0.309} & \textbf{0.220} & \textbf{0.343} & \textbf{0.485} &  & \textbf{0.241} & \textbf{0.159} & \textbf{0.265} & 0.398 &  & \textbf{0.339} & \textbf{0.238} & \textbf{0.395} & \textbf{0.528} \\
+ReaLitE & 0.285 & 0.197 & 0.319 & 0.461 &  & 0.239 & 0.157 & 0.265 & \textbf{0.400} &  & 0.333 & 0.232 & 0.387 & 0.524 \\
+\approach & 0.303 & 0.214 & 0.336 & 0.480 &  & 0.236 & 0.155 & 0.260 & 0.395 &  & 0.326 & 0.224 & 0.380 & 0.518 \\
\bottomrule
\end{tabular}
\end{table}

Table~\ref{tab:lp} reports link prediction performance for KGE models trained with and without co-training with \approach, and  ReaLitE \cite{klironomos2025realite} on FB15K-237, YAGO15K, and DB15K. As ReaLitE does not report results for all model-dataset pairs used in this study, following the paper’s recommendation of linear fusion for link prediction tasks, we ran multiple literal aggregation variants (learnable, min, and mode) and report the best scores.
Across all experiments, the underlying KGE models use the same embedding dimensionality for entities and relations; both ReaLitE and \approach only introduce additional literal-related parameters and use the same training configurations. Unlike ReaLitE's per-model-dataset hyperparameter optimisation, we evaluate every model under identical conditions without additional tuning.
The results show that \approach improves link prediction most consistently for bilinear models, with clear gains for Keci, DistMult, QMult, and OMult across datasets. For TransE and RotatE, co-training yields a small performance reduction, which we trace to gradient conflict and scoring-function constraints. Absolute performance remains modest across the 24 co-trained model-dataset settings in Table~\ref{tab:lp}: MRR ranges from 0.179 to 0.382 and Hits@1 from 0.115 to 0.299, and 18 of the 24 settings stay below an MRR of 0.300.
ReaLitE is more variable. It is competitive in selected cases, such as DualE on YAGO15K and QMult on DB15K, but also underperforms on some model-dataset pairs. Although ReaLitE outperforms \approach in some individual model-dataset cases, \approach still consistently improves over the corresponding base KGE variant for the bilinear models, indicating a more stable benefit from literal co-training. As mentioned in the original work ~\cite{klironomos2025realite}, ReaLitE's gains are dataset- and relation-dependent and not guaranteed across all configurations. Overall, even when co-training with \approach does not improve link prediction, the literal-aware evaluations show that it makes KGE models meaningfully encode and exploit literal information.

We further compared the literal prediction performance of the base models and their literal-augmented versions and report the results in the supplementary material. The results show that co-training improves literal prediction performance for most attributes, while reducing performance for some. Although KGE models with multiplicative scoring functions underperform translation-based models in literal prediction with \approach, combined training still improves their literal prediction performance. On YAGO15K, combined training reduces MAE for $diedOnDate$ in 8 of 8 models and for latitude in 5 of 8. On FB15K-237 the picture is mixed: $date\_of\_birth$ improves in 5 of 8 models, whereas $release\_date$, latitude, and longitude worsen in most or all of them. Note that the literal prediction performance discussed here for the literal augmented models refers to their performance via combined training, not post-literal-augmentation regression.

\subsubsection{Literal-Awareness Evaluation.} 
We use the methodology proposed by Blum et al.~\cite{blum2024numerical} to evaluate whether co-training KGE models with \approach makes their embeddings truly ``literal-aware''. The evaluation augments FB15K-237 with synthetic numerical attributes and attribute triples. For a subset of entities, values sampled from a uniform distribution determine a synthetic relation, $(e, r_{\text{syn}}, c_{\text{high}})$ if the value exceeds $0.5$ and $(e, r_{\text{syn}}, c_{\text{low}})$ otherwise. Models are trained on the augmented dataset and evaluated on whether they score $(e, r_{\text{syn}}, c_{\text{high}})$ above $(e, r_{\text{syn}}, c_{\text{low}})$ consistently with each entity's value. To test whether models use the numerical information itself, we repeat the evaluation with the values randomized but the graph structure unchanged, so a model relying on literal values should degrade under randomization, while one that performs similarly cannot distinguish meaningful from random values.

Table~\ref{tab:ablation} reports the accuracy of the models in classifying the synthetic relational triples under both settings. Models co-trained with \approach achieve higher accuracy on the original numerical values than on the randomized ones, while models augmented with LiteralE or ReaLitE perform equivalently in both settings, unable to distinguish actual from random values. This follows from how the approaches incorporate literal values: \approach augments entity embeddings via numerical reasoning, whereas LiteralE and ReaLitE incorporate literal values directly into the embeddings through a parametric transformation. Their link-prediction gains from literal augmentation may therefore stem from the additional model parameters rather than genuine reasoning over numerical values.
For evaluation settings and dataset construction, see the original work~\cite{blum2024numerical}.

\begin{table}[!t]
 \caption{Accuracy of models on classification of relational triples based on the augmented original and randomized values of the entities on the synthetic FB15K-237 dataset.}
    \centering
    \small
    \setlength{\tabcolsep}{1pt}
    \label{tab:ablation}
 \resizebox{\columnwidth}{!}{\begin{tabular}{lccccccccc}
\toprule
\textbf{Metric} & \textbf{Lit. Model} & \textbf{ComplEx} & \textbf{DistMult} & \textbf{DualE} & \textbf{Keci} & \textbf{OMult} & \textbf{QMult} & \textbf{RotatE} & \textbf{TransE}\\

\midrule
\textbf{Acc\_Rand} 
& LiteralE & 0.500 & 0.513 & 0.484 & 0.509 & 0.480 & 0.495 & 0.499 & 0.509 \\

& ReaLitE & 0.503 & 0.506 & 0.509 & 0.481 & 0.480 & 0.503 & 0.504 & 0.497 \\
& \approach & 0.499 & 0.512 & 0.474 & 0.479 & 0.504 & 0.521 & 0.498 & 0.499 \\

\midrule
\textbf{Acc\_Org} 
& LiteralE & 0.500 & 0.513 & 0.484 & 0.509 & 0.480 & 0.495 & 0.498 & 0.508\\
& ReaLitE & 0.501 & 0.495 & 0.509 & 0.496 & 0.524 & 0.504 & 0.501 & 0.498\\
& \approach & \textbf{0.635} & \textbf{0.711} & \textbf{0.793} & \textbf{0.693} & \textbf{0.827} & \textbf{0.813} & \textbf{0.860} & \textbf{0.816}\\

\bottomrule
\end{tabular}}
\end{table}

\section{Discussion}
\subsubsection{Literal Prediction and Augmentation.}

For literal prediction, training separate regression models per attribute performs worse than a single shared attribute-conditioned model, indicating that cross-attribute sharing provides useful regularization. At the same time, adding further layers yields only marginal gains, suggesting that regression complexity alone is not the bottleneck. A more fundamental constraint is that the regression model can only exploit information already encoded in the embeddings. Since KGE models are optimized for link prediction, entity embeddings primarily capture relational structure rather than semantic descriptions of entities, which limits literal prediction regardless of decoder capacity. This also motivates augmenting entity embeddings with attribute information directly. Our approach performs comparatively better with translation-based models such as TransE and RotatE, whose linearly constrained embedding geometries likely preserve neighbourhood information in a form our decoder can more readily exploit. It struggles with bilinear models such as DistMult and ComplEx, whose higher-order multiplicative interactions may not expose literal information in a linearly separable manner. A linear probe supports this: per-attribute ridge regressions on the frozen 64-dimensional FB15K-237 embeddings reach a mean $R^2$ of 0.28 for TransE against 0.18 for ComplEx and DistMult. As future work, we plan to investigate scoring-function-aware literal decoders that adapt to the underlying KGE geometry.

The link prediction gains under co-training show that incorporating numerical literals enriches entity embeddings with both relational and attribute information. These gains are concentrated in bilinear (multiplicative) models, which impose few constraints on entity placement and can therefore accommodate the literal signal without disrupting the structural objective. Aggregated over 4 bilinear models, 3 datasets, and 5 random seeds, this improvement is statistically significant. A one-sided Wilcoxon signed-rank test on the ranking metrics (Hits@\{1,3,10\}, MRR) confirms that co-training improves over KGE-only link prediction ($p < 0.001$).

Translational models show little co-training benefit, which we attribute primarily to their scoring-function constraints. These models enforce strict geometric relations (TransE requires $\mathbf{h} + \mathbf{r} \approx \mathbf{t}$, RotatE requires $\mathbf{h} \circ \mathbf{r} \approx \mathbf{t}$ in complex space), tightly coupling entity placement to relation structure and leaving the literal gradient little freedom to reposition embeddings without violating the structural objective. Measuring the cosine similarity between the KGE and literal gradients on shared embeddings in FB15k-237, we find that TransE produces conflicting gradient directions in 33\% of training steps, nearly double the 18\% of DistMult and Keci. The gradient-balancing mechanism counteracts this conflict, keeping performance close to the base KGE models. A secondary factor may contribute. Since TransE and RotatE already outperform bilinear models in literal prediction without co-training (Table~\ref{tab:kge_litpreds}), their geometry may already encode attribute-correlated structure, leaving less marginal signal for co-training to add.

\subsubsection{Probing Model Memorization.}
In existing literal prediction datasets, entities $e$ may appear in both training and test splits even though entity--attribute pairs $(e, a)$ do not, letting a model memorize attribute values via entity embeddings rather than generalize. To test this, we build $\kg'$ with a revised split over attribute triples only, leaving relation triples untouched. All attribute triples from the training, validation, and test sets are merged, and entities are assigned to train or test at a 70/30 ratio with a fixed random seed (42), enforcing $\entities'_{\text{train}} \cap \entities'_{\text{test}} = \emptyset$. This yields 16,248/7,012 attribute triples for FB15K-237-disjoint and 16,454/7,078 for YAGO15K-disjoint. Embeddings remain transductive over the full graph, so the split tests generalization to entities never used as supervised literal targets during training. It is not an inductive setting: \approach\ requires a trained embedding for every entity whose attributes it predicts. \approach\ performs comparably on this disjoint attribute split relative to the standard split, and better on some attributes, indicating it learns a transferable mapping from embedding space to values rather than memorizing per-entity targets.

\subsubsection{Performance under Reduced Datasets.}
We evaluate \approach\ on literal prediction under 80\%, 60\%, and 40\% subsamples of the training set of attribute triples (stratified by attribute), keeping the test set and pre-trained KGE embeddings fixed. \approach degrades gracefully, with most attributes within a few per cent $\Delta$MAE at 80\% and close to full-data accuracy at 60\%. Visible degradation appears only at 40\%, concentrated in date- and coordinate-valued attributes, though \textit{loc.area} and \textit{loc.date\_founded} change little at any ratio. The supplementary material reports the full per-attribute results.
\subsubsection{Limitation and Future Work.}
In this work, we focused only on predicting and augmenting numerical attributes in knowledge graphs. A limitation of the current approach is that it does not consider other types of literals, such as named individuals, boolean values, or textual descriptions of entities. Our experiments on link prediction employ an embedding dimensionality of 64 across knowledge graphs comprising between 12{,}000 and 15{,}000 entities, utilizing a fixed hyperparameter configuration for each respective experiment type. Whether the reported patterns hold at higher dimensionality, on larger graphs, or under model-specific tuning therefore remains untested. Extending the approach to handle these types of literals is an important direction for future work. Another promising direction is multi-hop query answering with literals. Integrating our approach under the LitCQD framework would enable state-of-the-art KGE models to answer multi-hop queries involving literals in a model-agnostic manner. Our approach also updates only the entity embeddings, so updating relation embeddings during joint training through the literal loss is another promising direction.

\section{Conclusion}
\label{section:conclusion}
In this work, we introduced a neural embedding regression model, dubbed \approach, to predict numerical attributes of entities in knowledge graphs by leveraging pre-trained embeddings. Experiments across datasets show that \approach is competitive with state-of-the-art numerical attribute prediction methods: \approach\ or \approachbase achieves the best or second-best MAE on 8 of 11 FB15K-237 attributes and on all 7 YAGO15K attributes. We also repurpose the existing KGs DB15K and Mutagenesis as benchmarks for numerical attribute prediction, reporting the first results on both, where \approach\ or \approachbase ranks best or second-best on all 9 DB15K and all 4 Mutagenesis attributes. We also designed a co-training framework that enables embedding methods to incorporate and predict numerical attributes in KGs simultaneously. Experiments across datasets suggest that our approach improves link prediction for embedding models with multiplicative scoring functions while also making them literal-aware.

\begin{credits}
\subsubsection{\ackname}
This work has been supported by the Ministry of Culture and Science of North Rhine-Westphalia (MKW NRW) within the project SAIL under the grant no NW21-059D, by  the project ``WHALE'' (LFN 1-04) funded under the Lamarr Fellow Network programme by the Ministry of Culture and Science of North Rhine-Westphalia (MKW NRW), and by the German Federal Ministry of Research, Technology and Space (BMFTR) within the project KI-Akademie OWL under the grant no 16IS24057B.

\subsubsection{\discintname}
The authors have no competing interests to declare that are relevant to the content of this article.
\end{credits}

\bibliographystyle{splncs04}
\bibliography{bibliography}

\end{document}